\documentclass[conference]{IEEEtran}

\usepackage{graphicx}
\usepackage{caption}
\DeclareCaptionFormat{citation}{%
   \ifx\captioncitation\relax\relax\else
     \captioncitation\par
   \fi
   #1#2#3\par}
\newcommand*\setcaptioncitation[1]{\def\captioncitation{\textit{Source:}~#1}}
\let\captioncitation\relax
\usepackage{mwe}

\IEEEoverridecommandlockouts
\usepackage{cite}
\usepackage{amsmath,amssymb,amsfonts}
\usepackage{algorithm}
\usepackage{algorithmic}
\usepackage{graphicx}
\usepackage{textcomp}
\usepackage{xcolor}
\def\BibTeX{{\rm B\kern-.05em{\sc i\kern-.025em b}\kern-.08em
    T\kern-.1667em\lower.7ex\hbox{E}\kern-.125emX}}
\usepackage[colorlinks,citecolor=red,urlcolor=blue,bookmarks=false,hypertexnames=true]{hyperref} 

\begin{document}

\title{Evaluation of Time Series Forecasting Models for Estimation of PM2.5 Levels in Air \\
{\footnotesize \textsuperscript{*}Note: This paper is accepted and presented in the 6th I2CT 2021 conference. The final version of this paper will appear in the conference proceedings.}
}
\author{\IEEEauthorblockN{Satvik Garg\textsuperscript{1},
Himanshu Jindal\textsuperscript{2}}
\IEEEauthorblockA{Department of Computer Science,
Jaypee University of Information Technology\\
Solan\\
Email: \textsuperscript{1}satvikgarg27@gmail.com,
\textsuperscript{2}himanshu.jindal@juitsolan.in}}

\IEEEoverridecommandlockouts
\IEEEpubid{\makebox[\columnwidth]{978-1-7281-8876-8/21/\$31.00~\copyright{}2021 IEEE \hfill} \hspace{\columnsep}\makebox[\columnwidth]{ }}
\maketitle
\begin{abstract}
Air contamination in urban areas has risen consistently over the past few years. Due to expanding industrialization and increasing concentration of toxic gases in the climate, the air is getting more poisonous step by step at an alarming rate. Since the arrival of the Coronavirus pandemic, it is getting more critical to lessen air contamination to reduce its impact. The specialists and environmentalists are making a valiant effort to gauge air contamination levels. However, it's genuinely unpredictable to mimic sub-atomic communication in the air, which brings about off-base outcomes. There has been an ascent in using machine learning and deep learning models to foresee the results on time series data. This study adopts ARIMA, FBProphet, and deep learning models such as LSTM, 1D-CNN, to estimate the concentration of PM2.5 in the environment. Our predicted results convey that all adopted methods give comparative outcomes in terms of average root mean squared error. However, the LSTM outperforms all other models with reference to mean absolute percentage error.
\end{abstract}


\section{Introduction}

From the smog looming over metropolitan regions to pollution inside the home, air defilement represents a huge threat to well-being and the atmosphere. Air contamination accounts for an expected 4.2 million deaths per year because of stroke, coronary illness, cellular breakdown in the lungs, intense and constant respiratory sickness [1].

Air toxins can be present in the air anyplace, whether it is inside or outside. 
It incorporates gaseous toxins, for example, Carbon Monoxide (CO), Sulfur Dioxide (SO2), Nitrogen Dioxide (NO2), Ozone(O3), and particle matters like PM2.5, PM10. The health impacts from particulate matter are controlled by its size, composition, source, solubility, and capacity to create reactive oxygen. PM2.5 represents air particulate matter with a diameter under 2.5 micrometers, represents 3\% the diameter of human hair [2].

An examination was published by the Journal of American Medical Association to evaluate the connection between long-term exposure to fine particulate air contamination with the cellular breakdown in the lungs and cardiopulmonary mortality. It recommends that for every 10-microgram/m3 increase in particulate air contamination, the risk raised by an average of 4-8\% of heart stroke and lung cancer mortality [3].

An early assessment of air contamination levels encourages the policymakers to choose the time specific strategies that shall need to be executed for supporting the residents, for example, sponsoring the utilization of public transport, offering free defensive facial covers, and financing clinical tests for asthma patients so that they can plan monetarily and successfully. Notwithstanding, these advantages and strategies are dependent upon knowing pollution levels in advance.

The forecasting of time-related data is a difficult problem because of the unknown changes in air contamination level patterns and conditions. In this research, we explored for finding the solutions that offered the best outcomes concerning lower prediction errors.  In this regard, we used the stochastic model ARIMA [4], additive model FBProphet [5], and deep learning models LSTM [6], 1D-CNN [7].

This research work is categorized as follows: Section
II provides the literature survey related to time series analysis. Section III describes the framework adopted in this work for forecasting. The analysis and evaluation of predictions using various metrics is presented in Section IV. Section V concludes the paper.

\section{Literature Survey}

Forecasting of time series data is well known and excellent choice for analyzing the economy, stocks, human activity data, traffic, climate, sales, social media mining, and much more. Time series forecasting analyzes lags known as path observation to gain useful features from data to forecast future values using past data. 

In early times, time-series data was forecasted using standard regressive models like AR, MA, ARMA, ARIMA [8]. These models are quite usual in forecasting economic and financial data. Still, they had some limitations as the models were not meant to analyze the non-linear behavior between variables and also when the data exhibit conditional covariance implies change in variance over time. However, one could solve this problem by integrating it with the Generalized Auto-regressive Conditional Heteroskedasticity (GARCH), but it's quite difficult to optimize its parameters [9]. 

Weitao Wang et al. [11] proposed a simple and effective hybrid model for forecasting traffic flow. This study finds out that traffic flow analysis is subjective to both linear and non-linear relationships of data. The proposed hybrid model consists of the ARIMA model for linear fitting and Radial basis function artificial neural network (RBF-ANN) for the non-linear fitting of data.

Examining the requirement for fast and compelling techniques for performing web forecast of network traffic, Hao Yin et al. [12],  proposed an adaptive autoregression (AAR) model. The idea was to integrate an adaptive order-selection and memory shortening technique to uphold the online forecast of dynamic network traffic data. The model achieved good exactness and low computation cost dissimilar to conventional Box–Jenkins time arrangement models like AR, MA, ARMA, ARIMA, etc.

The methods in deep learning was used and developed to deliver the difficulties identifying with the forecasting models. Considering the significance of CNN in various fields, Zhaoyi Xu et al. [13], adopted CNN for forecasting stock indexes, and the impacts of chronicled factors on the model were dissected. Finally, a couple of stock indexes were anticipated to confirm legitimacy and viability of the proposed model. The author also generated a hybrid model combined with CNN, which further improved the CNN network model.

The study [14] examines the use of deep neural networks; gradient boosted trees, random forest, and a simple ensemble of the models to forecast the S\&P 500. The authors reported that random forest achieved better results than gradient boosted trees and deep neural networks.

Recently, the Facebook data science team deployed an open-source additive time series model called FBProphet. Assessing the open-source algorithm, faster results, and accuracy, Alabi et al. [15] used FBProphet to estimate COVID-19 passings and confirmed cases. The accuracy achieved by prophet was 79.6\% for the data from World Health Organization.

From the survey, we aim to show the time series analysis from basic models like ARIMA to recently developed models like FBProphet. We also discussed deep learning approaches like CNN, and machine learning techniques like the random forest, gradient boosting. The problems were also reviewed using the ARIMA model and how one can solve them using GARCH. Generally, all studies focused on creating a slight change in the existing model and modify them to provide outcomes. However, a proper comparison was somehow limited, covering various approaches for evaluation provided in this research.

\section{Methodologies}
The dataset used in this research is Beijing Multi-Site Air-Quality, taken from the UCI Machine Learning Repository [16]. It incorporates hourly air contamination information from 12 broadly controlled air-quality observing locales (stations) between the timeframe March 1st, 2013 to February 28th, 2017. It contains 18 features, which are given in Fig. 1. 
The target variable is pollution measured in PM2.5. For each station, the dimensions are 18 by 35064, so for a total of 12 stations, it is equivalent to 18 by 420768.
\begin{figure}[htbp]
\centerline{\includegraphics[width=8cm, height=5cm]{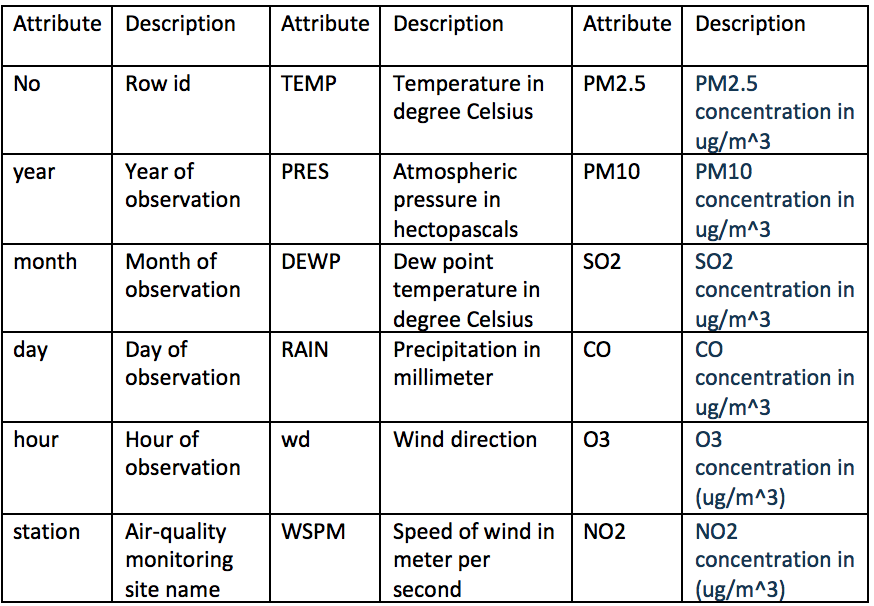}}
\caption{Attribute Information.}
\label{fig}
\end{figure}

Fig. 2 explains and describes the framework adopted in this research for measuring the PM2.5 levels. We divided the framework into three phases, namely, Data Preprocessing, Modeling, and Evaluation phase.

\begin{figure}[htbp]
\centerline{\includegraphics[width=6.5cm, height=6.5cm]{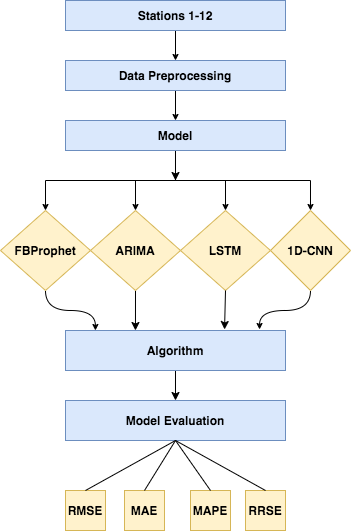}}
\caption{Flowchart of the framework adopted.}
\label{fig}
\end{figure}

\subsection{Data Preprocessing}
Prior training the models it is extremely important to clean the data. We used some basic data preprocessing techniques in this work. Instead of dropping rows containing null values, we applied the 'fillna' method using pandas library [17] to forward fill the empty values in the dataset. The 'wd' feature was converted from categorical to numerical values using the label encoder function from the sci-kit learn toolkit [18]. We took an hour, day, month, year attributed from the data and formed an ordered day by day level 'DateTime' column, which makes the total row count equals 1421 for each station. The data was then divided into a training and a test set. The training data consists from March 1st, 2013 to February 28th, 2016, which accounts for 75\% of data and test data from March 1st, 2016 to February 28th, 2017 (25\%).

For deep learning methods, LSTM, 1D-CNN, considering the necessity for validation data [19], 20 percent of training data was used for validation purposes.

In fundamental terms, there is a stationary (uniform) and non-stationary (non-uniform) time series. A uniform time series is one whose quantifiable properties such as the mean, variance, and autocorrelation, are congruent with time. Henceforth, a non-uniform time series represents a change in properties over time which shows the presence of a trend.
Non-uniform time series should be first changed over into detrended arrangement prior to applying the models. If the time arrangement is dependably extending or diminishing over time, the example mean and the variance will form with the size of the example, and this will reliably deprecate the mean and variance in future periods.

To examine in-case the time series is uniform or not, Augmented Dickey-Fuller (ADF) test [20] is used. Prior going to the ADF test, we should initially comprehend, 'what is the Dickey-Fuller test.' It is a unit root test that checks for the null hypothesis ($\alpha{=1}$) in the equation below:
\begin{equation}
    y_t = c + \beta{t} + \alpha {y_{t-1}} + \phi\Delta Y_{t-1} + e_t
\end{equation}
where,\\
${y_{t-1}}$ = first Lag of time series \\
$\Delta Y_{t-1}$ = first difference of the series at time (t-1)\\
In simple words, if the value of a equals 1, it indicates the presence of unit root. Thereby, the series has taken to be non-stationary.
The ADF test allows higher-order regressive processes in the model by including $\Delta Y_{t-p}$.
\begin{equation}
    y_t = c + \beta{t} + \alpha {y_{t-1}} + \phi_{1}\Delta Y_{t-1} +\phi_{2}\Delta Y_{t-2}+\dotsm +\phi_{p}\Delta Y_{t-p}+ e_t
\end{equation}
The null hypothesis is similar to the dickey fuller test. However, the p-value acquired ought to be less than the significance level of 5\% to dismiss the null hypothesis that is the presence of unit root.
We applied this test on our target variable PM2.5 and discovered that our series is stationary.
Figures 3 and 4 show the subplots of six main air pollutants and six external relevant meteorological variables on Aotizhongxin station, respectively. One can clearly see that there is a seasonality present in various features like Temperature, Dew point Temperature, Precipitation, CO, NO2, O3 and SO2.

\begin{figure}[htbp]
\centerline{\includegraphics[width=8cm, height=8cm]{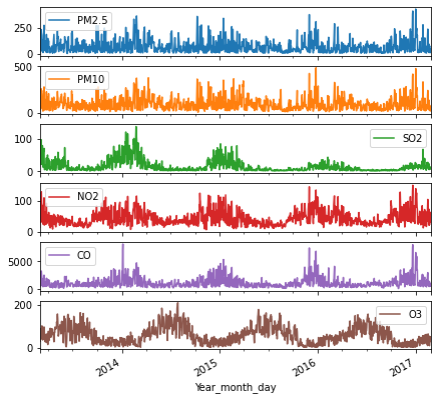}}
\caption{Plot of 6 main atmospheric pollutants concentration with respect to days on station Aotizhongxin.}
\end{figure}
\begin{figure}[htbp]
\centerline{\includegraphics[width=8cm, height=8cm]{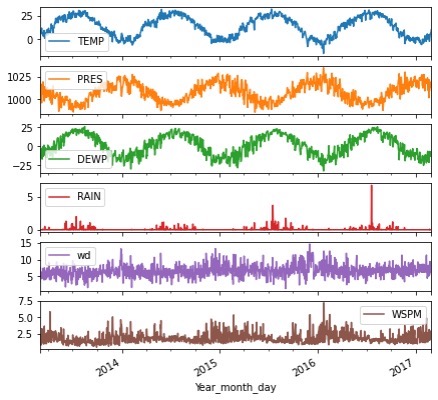}}
\caption{Plot of 6 meteorological factors with respect to time on station Aotizhongxin}
\end{figure}

\subsection{Modeling}
Four different types of models, namely, FBProphet, ARIMA, LSTM, 1D-CNN has been adopted in this work to evaluate time series forecasting PM2.5 levels.
\subsubsection{FBProphet}
Facebook prophet [5] is an open-source tool developed by Facebook used for time series analysis derive from a decomposable additive model. It takes into account holidays, and it usually fits nonlinear data with yearly, weekly and daily seasonality. Prophet uses time as a regressor and fits various linear and nonlinear functions of time as components, which is combined given in this equation:
\begin{equation}
y(t) = g(t) + s(t) + h(t) + e(t)
\end{equation}
where,\\
$y(t)$: predictions (forecast).\\
$g(t)$: trend alludes to changes throughout a significant stretch of time \\
$s(t)$: refers to seasonality for example, weekly, daily, yearly.\\
$h(t)$: holidays\\
$e(t)$: error term represents any surprising changes not obliged by the model\\
The Facebook prophet has multiple trend, seasonality, changepoints, and holiday parameters which needs to be tuned for better results. Parameters and its values that optimized in this research is shown in Fig. 5. 
Changepoint refers to those points at which time series experience an abrupt change. These progressions can be because of anything, for instance, unexpected catastrophe, new government guidelines, and so forth. 
The parameter optimization resulted in 144 possible model counts, assessed by evaluation metrics.

\begin{figure}[htbp]
\centerline{\includegraphics[width=8cm, height=2.4cm]{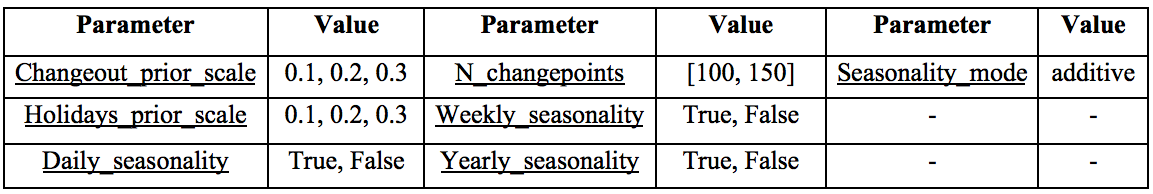}}
\caption{Hyperparameters FBProphet}

\end{figure}

\subsubsection{ARIMA}
ARIMA(p,d,q) [4], composition of autoregression, integrated and moving average is a regressive model used to forecast time series data 
where p, d, q referred as autoregression order. 
Auto Regression, AR(p), is a part of the ARIMA model based on the idea that it uses its own lags (past values) as predictors, where p is a boundary of the number of lags taken in
\begin{equation}
    y_t = \alpha + \displaystyle\sum\limits_{i=0}^p\beta_{i}Y_{t-i} + \epsilon_{t}
\end{equation}
where, $\alpha$ represents intercept, $\epsilon_{t}$ adures white noise , $\displaystyle\sum\limits_{i=0}^p\beta_{i} $ are the coefficients of the past values (lags) given by $\displaystyle\sum\limits_{i=0}^p Y_{t-i}$ which are calculated by the model.\\
Moving Average, MA(q), utilizes residual error of past time points to foresee current and future predictions. Moving normal (MA) eliminates arbitrary developments from a time series. The parameter q is the number of lags forecast errors that utilized to compute current values.

\begin{equation}
    y_t = \alpha + \displaystyle\sum\limits_{i=0}^q\phi_{i}\epsilon_{t-i} + \epsilon_{t}
\end{equation}
$\displaystyle\sum\limits_{i=0}^q\epsilon_{t-i}$ represent error terms of the respective lags.\\
Integrated, I(d), property helps make time-series data stationary to eliminate time dependency and trend. Parameter 'd' represents the degree of difference, which means the number of times the data was differenced. If a time series is stationary, then its degree of difference is zero.\\
The ARIMA(p,d,q) model is described as:
\begin{equation}
    y_t = \alpha + \displaystyle\sum\limits_{i=0}^p\beta_{i}Y_{t-i} + \displaystyle\sum\limits_{i=0}^q\phi_{i}\epsilon_{t-i} + \epsilon_{t}
\end{equation}
The standard ARIMA models manually take input values (p,d,q) using autocorrelation, and various other statistical tests. In this study, we use Auto Arima from the pmdarima package [10], which automatically evaluates p, d, q values on its own. Considering our time series is stationary, we kept d=0 and attempted with the p, q values ranging from 0 to 5 to optimize the ideal value for the model.

\subsubsection{LSTM}
LSTM [6] represents long short term memory networks. It is a model or design that expands the short-term memory of recurrent neural networks that help determine the time series problems effectively. RNN deals with the current input by considering only the previous yield (feedback) and putting it in its memory for a brief timeframe (short-term memory). Thus, neglecting to store data for a long time implies the inefficiency of RNNs for dealing with long-term dependencies. Different issues with RNNs are vanishing and exploding gradients problems during the training phase by backtracking. This would halt network from learning since the updated weights become smaller and smaller or bigger and bigger. The network should be rebuilt in such a way so that it scales down the scaling factor to one. It should be possible by utilizing different gate units in memory blocks, associated through layers, as appeared in Fig. 6 and called LSTM. This research utilizes just a single LSTM layer with only 128 neurons in it, followed by a fully connected layer for prediction.

\begin{figure}[h]
\centerline{\includegraphics[width=8.6cm, height=4cm]{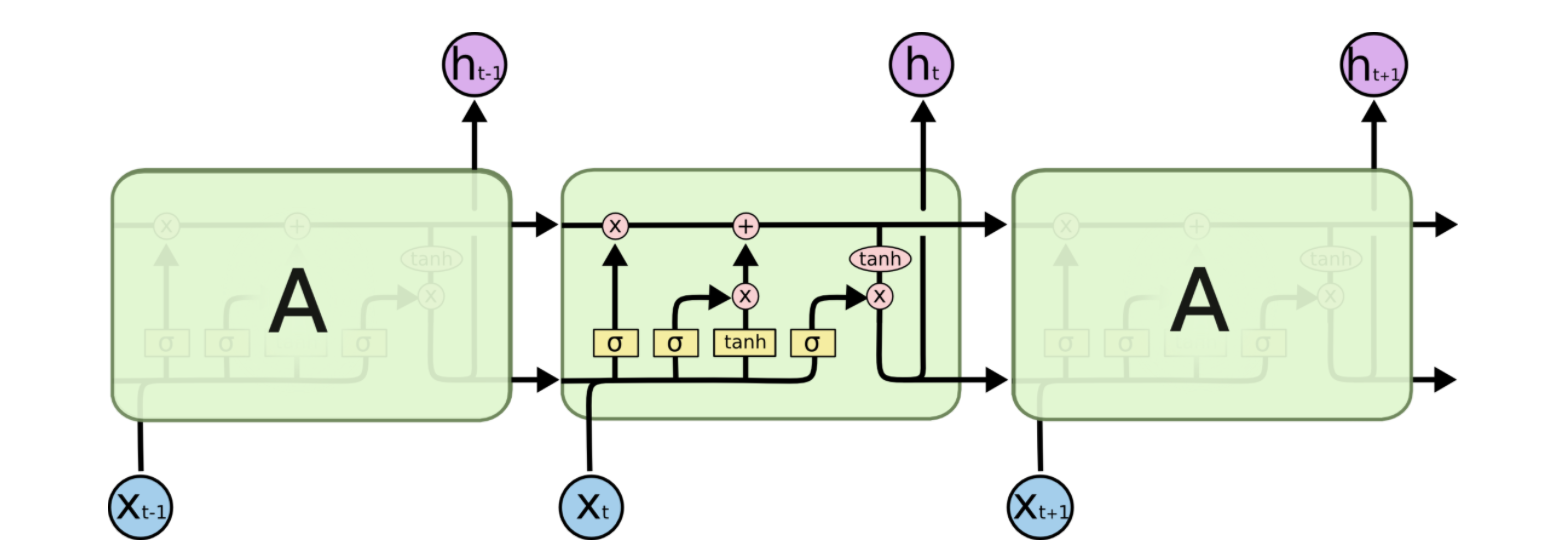}}
\setcaptioncitation{\url{http://colah.github.io/posts/2015-08-Understanding-LSTMs/}}
\caption{LSTM repeating network layer}

\end{figure}

\subsubsection{CNN}
Convolutional neural network models [7] were mainly produced for picture classification problems. The convolutional layers extract the features from a two-dimensional input, referred to as feature engineering. This equivalent process could outfit on one-dimensional arrangements of data, for example, sequence problems. The advantage to implement CNNs for sequence data is that they can learn from the raw time series data straightforwardly as it provides an architecture to perform smoothing parameters. The underlying two layers of CNN are typically a convolutional layer and a pooling layer. Both perform smoothing, followed by a Dense layer to use the smoothed data, and perform well on a forecasting task. A flatten layer is also used between the max-pooling layer and the fully connected layer to reduce generated multidimensional features to a one-dimensional vector. 

\subsection{Algorithms}
As shown in Algorithm 1 and 2, input to the proposed algorithm is PM2.5 series (Data) and Features series set of 12 stations.
Features represent external variables like NO2, PM10, SO2, CO, wd, DEWP, and many more. These features help to build the model efficiently. After building the model, both algorithms output the results of models using four error metrics, RMSE, MAE, MAPE, RRSE. The first step is to divide the dataset for training and testing purposes. Algorithm 2 followed a similar train test split like Algorithm 1 (75:25). However, the 20\% of training data and features taken as validation data appeared in lines 7 and 10. For deep learning methods, it is good practice to normalize the data to the range of 0 to 1 since the networks are sensitive to the scale of input data. The MinMaxScaler preprocessing function is used as appeared in lines 1-4. 

During the model fitting, for FBProphet as shown in Algorithm 1, applied various hyperparameters that were shown in Fig. 5 using a simple loop. The model was then fitted using a fit() method in line 10. Each fitted model predicted the test data using Test Features as given in line 11. 

ARIMA followed the similar approach like FBProphet. However, we didn't use any loop as the values of hyperparameters are generated automatically. The auto\_arima method in line 14 will try various hyperparameters on its own and returns the best model having the lowest Akaike information criterion (AIC) score [22].

\begin{algorithm}
\caption{Algorithm for FBProphet and ARIMA}
  \algsetup{linenosize=\tiny}
  \scriptsize
 \begin{algorithmic}[1]

 \renewcommand{\algorithmicrequire}{\textbf{Input:}}
 \renewcommand{\algorithmicensure}{\textbf{Output:}}
 \REQUIRE Data, Features
 \ENSURE  Evaluation metrics of the predicted data
 \\ \textit{\textbf{Data split}} : \textit{75\% train and 25\% test data}\\
  \textit{\textbf{Train}}\\
  \STATE {$count$ $\leftarrow$ $length(Data)*0.75$}\\
  \STATE {$X$ $\leftarrow$ $Data(0:count)$}\\
  \STATE {$Z$ $\leftarrow$ $Features($0 : $count)$}\\
  \textit{\textbf{Test}}\\
  \STATE {$x$ $\leftarrow$ $Data(count$ : $)$}\\
  \STATE {$z$ $\leftarrow$ $Features(count$ : $)$}\\
  \textit{\textbf{Model fitting FBProphet}}
  \STATE {$parameters$ $\leftarrow$ $Hyperparameters$}
  \FOR {each $p$ in $parameters$}
  \STATE {$model$ $\leftarrow$ $Prophet(p, interval\_width = 0.95)$}
  \STATE {$model.add\_regressor(Z)$}  
  \STATE {$model.fit(X)$}
  \STATE {$forecast$ $\leftarrow$ $model.predict(z)$}
  \RETURN $rmse$, $mae$, $mape$, $rrse$ 
  \ENDFOR\\
  \textit{\textbf{Model fitting ARIMA}}
  \STATE {$model$ $\leftarrow$ $auto\_arima(X,exogenous$=$Z)$}\\
  \STATE {$model.fit()$}\\
  \STATE {$forecast$ $\leftarrow$ $model.predict(n\_periods=len(x),z)$}\\
  \RETURN $rmse$, $mae$, $mape$, $rrse$ 
  \end{algorithmic} 
\end{algorithm}

As shown in Algorithm 2, the deep learning methods, LSTM, 1-D CNN, trained independently for 200, 400, 600, 800, 1000 epochs to check for underfitting and overfitting. However, for LSTM additionally applied the 'tanh' and 'relu' activation function to check for the nonlinearity of 'relu' to help in improving the model. These parameters resulted in the formation of two loops which made a sum of ten distinct models per station for LSTM. 1-D CNN alike to LSTM except there is only one loop used for epochs hyperparameter resulting in five models per station. We use 'adam' optimizer for model compilation as appear in lines 20 and 34.

\begin{algorithm}
\begin{footnotesize}
  \algsetup{linenosize=\tiny}
  \scriptsize
\caption{Algorithm for LSTM and 1D-CNN}
 \begin{algorithmic}[1]

 \renewcommand{\algorithmicrequire}{\textbf{Input:}}
 \renewcommand{\algorithmicensure}{\textbf{Output:}}
 \REQUIRE Data, Features
 \ENSURE  Evaluation metrics of the predicted data
 \\ \textit{\textbf{Data split}} : \textit{75\% training + validation and 25\% test data}\\
 \textit{\textbf{Normalization}}
 \STATE {$object1$ $\leftarrow$ $MinMaxScaler()$}
 \STATE {$object2$ $\leftarrow$ $MinMaxScaler()$}
 \STATE {$Data2$ $\leftarrow$ $object1.fit\_transform(Data)$}
 \STATE {$Features2$ $\leftarrow$ $object2.fit\_transform(Features)$}\\
  \textit{\textbf{Train and Validation}}
  \STATE {$count$ $\leftarrow$ $length(Data2)*0.75$}
  \STATE {$X$ $\leftarrow$ $Data2(0:count)$}
  \STATE {$V$ $\leftarrow$ $X*0.20$}
  \STATE {$X1$ $\leftarrow$ $X-V$}
  \STATE {$Z$ $\leftarrow$ $Features2($0 : $count)$}
  \STATE {$v$ $\leftarrow$ $Z*0.20$}
  \STATE {$Z1$ $\leftarrow$ $Z-v$}\\
  \textit{\textbf{Test}}
  \STATE {$x$ $\leftarrow$ $Data(count$ : $)$}
  \STATE {$z$ $\leftarrow$ $Features(count$ : $)$}\\
  \textit{\textbf{Hyperparameters}}
  \STATE {$epochs$ $\leftarrow$ $[200,400,600,800,1000]$}
  \STATE {$activation$ $\leftarrow$ $['tanh', 'relu']$}\\
  \textit{\textbf{Model fitting LSTM}}
  \FOR {each $i$ in $activation$}
  \STATE {$model$ $\leftarrow$ $Sequential()$}
  \STATE {$model.add(LSTM(128),activation=i$)}  
  \STATE {$model.add(Dense(1))$}
  \STATE {$model.compile(loss = 'mae', optimizer = 'adam')$}
  \FOR {each $j$ in $epochs$}
  \STATE {$model.fit(Z1, X1, validation = (v,V), epochs = j)$}
  \STATE {$forecast$ $\leftarrow$ $model.predict(z)$}
  \STATE {$forecast$ $\leftarrow$ $object1.inverse\_transform(forecast)$}
  \RETURN $rmse,$ $mae,$ $mape,$ $rrse$ 
  \ENDFOR
  \ENDFOR\\
  \textit{\textbf{Model fitting 1D-CNN}}
  \STATE {$model$ $\leftarrow$ $Sequential()$}
  \STATE {$model.add(Conv1D(filters = 128, kernel\_size = 2,         activation = 'relu'))$} 
  \STATE {$model.add(MaxPooling1D(pool\_size = 2))$}
  \STATE {$model.add(Flatten())$}
  \STATE {$model.add(Dense(64, activation = 'relu'))$}
  \STATE {$model.add(Dense(1))$}
  \STATE {$model.compile(loss = 'mae', optimizer = 'adam')$}
  \FOR {each $i$ in $epocs$}
  \STATE {$model.fit(Z1, X1, validation\_data = (v,V), epocs = i)$}
  \STATE {$forecast$ $\leftarrow$ $model.predict(z)$}
  \STATE {$forecast$ $\leftarrow$ $object1.inverse\_transform(forecast)$}
  \RETURN $rmse,$ $mae,$ $mape,$ $rrse$ 
  \ENDFOR
\end{algorithmic} 
\end{footnotesize}
\end{algorithm}

\subsection{Metrics}
Let the letter be, $RMSE$ : root mean squared error, $MAE$ : mean absolute error, $MAPE$ : mean absolute percentage error, $RRSE$ : root relative squared error, $x_{i}$ : actual data, $y_{i}$ : predicted data, $n$ : length of test data.
The following metrics are used in this research to analyze the forecasted results:
\begin{equation}
    MAPE = (\frac{100}{n})\sum_{i=0}^n \frac{\left | x_{i} - y_{i} \right |}{\left | x_{i}  \right |}
\end{equation}
\begin{equation}
    RMSE = \sqrt{\frac{1}{n}\sum_{i=1}^{n}{(x_{i}-y_{i})^2}}
\end{equation}
\begin{equation}
    MAE = (\frac{1}{n})\sum_{i=1}^{n}\left | (x_{i} - y_{i} \right)|
\end{equation}
\begin{equation}
    RRSE = \sqrt{\frac{\sum_{i=1}^{n}(y_{i}-x_{i})^2}{\sum_{i=1}^{n}(x-x_{i})^2}} 
    \quad where\enspace x = (\frac{1}{n})\sum_{i=1}^{n}(x_{i})
\end{equation}

\section{Results}
In this evaluation, four distinct models were assessed, to be specific LSTM, CNN, ARIMA, FBProphet for the time Series examination on gauging PM2.5 levels on 12 Stations. We utilized test information for forecasts that represent 25 percent of the complete dataset. For deep learning methods, LSTM, CNN, 20\% of the training data adopted for validation purposes. RMSE, RRSE, MAPE, MAE scores were used to assess model execution. 

MAE is simply the mean of absolute error. MAE calculates errors on a similar scale which infers that it treats big and small errors equally. It isn't sufficient to investigate the forecasts appropriately. MAPE is sort of standardized absolute error, which permits the errors to be contrasted across data with various scales. MAPE is processed over each information point and averaged, and accordingly catches more mistakes and exceptions. It is also helpful to punish negative errors as the estimation of the actual data would be more smaller than forecasted data. On the other hand, RMSE is helpful for penalizing larger errors as the errors are squared which comparatively gives higher weight to larger values. RRSE simply measures the relativeness of the predictions with the average of actual values.

Table I and II shows the forecasting metrics results for examining PM2.5 levels on all 12 stations. The order for ARIMA is referenced alongside with the results to bring about in the tables. From the outcomes, one can without much of a stretch see that LSTM has better execution compared to different models for all evaluation metrics. LSTM has performed well in all stations yet for some stations like, Dingling, Dongsi, Gucheng, and Wanliu, the RMSE values are somewhat higher than ARIMA and FBProphet. 
For station Dingling, the RMSE and RRSE values are on the higher side compared to different stations.  The best MAPE score is 17.9, which is for Dongsi station by LSTM. Best RMSE and MAE scores are 16.6 and 11.5, respectively, for station Tiantan by LSTM. The best RRSE score is 0.23 for station Tiantan and Wanshouxigong by LSTM and CNN. 

\begin{table}[htbp]
\caption{Predicted Results on 12 stations}
\begin{center}
\begin{tabular}{|p{1.6cm}|p{1.6cm}|p{0.65cm}|p{0.65cm}|p{0.65cm}|p{0.65cm}|}
\hline
\textbf{Station }
&\textbf{Model}& \textbf{RMSE}& \textbf{MAE}& \textbf{MAPE}
& \textbf{RRSE}\\
\hline
& FBProphet& 20.1&13.2 &29.8&0.27 \\
Aotizhongxin& ARIMA(1,0,3)& 21.0&14.2 &33.4&0.29 \\
& LSTM& 19.2&12.4 &21.2&0.26 \\
& CNN& 25.0&15.9 &25.5&0.34 \\
\hline
& FBProphet& 18.9&13.2 &37.0&0.30 \\
Changping& ARIMA(2,0,0)& 19.2&13.3 &34.8&0.30 \\
& LSTM& 18.8&12.7 &28.0&0.30 \\
& CNN& 20.2&14.0 &37.1&0.32 \\
\hline
& FBProphet& 35.6&16.5 &45.3&0.53 \\
Dingling& ARIMA(2,0,0)& 35.1&16.2 &34.8&0.52 \\
& LSTM& 35.6&15.2&22.8&0.53 \\
& CNN& 36.7&16.4 &29.5&0.54 \\
\hline
& FBProphet& 20.5&15.1 &33.2&0.26 \\
Dongsi& ARIMA(1,0,0)& 19.8&14.1 &27.5&0.25 \\
& LSTM& 21.3&13.2 &17.9&0.27 \\
& CNN& 22.0&16.1 &26.9&0.28 \\
\hline
& FBProphet& 20.2&15.0 &34.5&0.27 \\
Guanyuan& ARIMA(2,0,0)& 20.2&14.7 &31.5&0.27 \\
& LSTM& 18.8&12.5 &20.3&0.25 \\
& CNN& 21.3&14.2 &25.6&0.28 \\
\hline
& FBProphet& 20.7&15.0 &36.2&0.27 \\
Gucheng& ARIMA(1,0,0)& 20.6&14.6 &31.2&0.27 \\
& LSTM& 22.2&15.2 &23.9&0.29 \\
& CNN& 23.2&14.3 &22.5&0.30 \\
\hline
\end{tabular}
\end{center}
\end{table}

It has been observed that aside from station Dingling, the distinction in the RMSE and MAE estimations for all stations is near 6 units for each of the four models. It shows that there is some variation present in the value of the errors and extremely huge errors are not likely to have happened. Notwithstanding, for station Dingling the difference is 16 units which recommends the presence of enormous mistakes which is being punished by RMSE. The more prominent the distinction between the RMSE and MAE esteems, the bigger the change in the individual mistakes in the sample.

\begin{table}[htbp]
\caption{Predicted Results on 12 stations}
\begin{center}
\begin{tabular}{|p{1.6cm}|p{1.5cm}|p{0.65cm}|p{0.65cm}|p{0.65cm}|p{0.65cm}|}
\hline
\textbf{Station }
&\textbf{Model}& \textbf{RMSE}& \textbf{MAE}& \textbf{MAPE}
& \textbf{RRSE}\\
\hline
& FBProphet& 18.5&12.2 &35.3&0.30 \\
Huairou& ARIMA(1,0,1)& 19.9&12.3 &27.6&0.33 \\
& LSTM& 17.2&11.9 &32.5&0.29 \\
& CNN& 18.9&14.0 &38.7&0.31 \\
\hline
& FBProphet& 20.7&15.3 &35.4&0.27 \\
Nongzhanguan& ARIMA(1,0,1)& 20.3&14.4 &29.4&0.27 \\
& LSTM& 18.9&13.2 &20.3&0.25 \\
& CNN& 19.8&14.0 &21.7&0.26 \\
\hline
& FBProphet& 19.8&13.9 &32.6&0.28 \\
Shunyi& ARIMA(4,0,0)& 19.9&13.9 &29.0&0.28 \\
& LSTM& 19.8&12.6 &23.9&0.28 \\
& CNN& 22.2&14.0 &25.6&0.31 \\
\hline
& FBProphet& 20.6&14.7 &32.6&0.28 \\
Tiantan& ARIMA(1,0,0)& 19.6&13.4 &26.4&0.27 \\
& LSTM& 16.6&11.5 &19.6&0.23 \\
& CNN& 18.4&12.5 &19.4&0.25 \\
\hline
& FBProphet& 19.6&14.7 &38.3&0.27 \\
Wanliu& ARIMA(1,0,0)& 19.3&14.2 &34.0&0.27 \\
& LSTM& 23.9&15.9 &23.3&0.33 \\
& CNN& 23.4&13.9 &22.6&0.33 \\
\hline
& FBProphet&19.3&13.9 &26.4&0.24  \\
Wanshouxigong& ARIMA(1,0,0)& 19.5&14.0 &26.4&0.25 \\
& LSTM& 17.8& 12.5&18.7&0.23 \\
& CNN& 18.3&12.8 &21.1&0.23 \\
\hline

\end{tabular}

\end{center}
\end{table}

The average errors of all stations are given in Table III to assess overall model performance. The deep learning based model, LSTM, CNN, outperforms ARIMA, and FBProphet in terms of mean absolute percentage error. At the same time, LSTM has the lowest error in all assessment metrics compared with other models. The MAPE value is highest for FBProphet, which is 34.7.
\begin{table}[htbp]
\caption{Averaged predicted Results}
\begin{center}
\begin{tabular}{|c|c|c|c|c|}
\hline
\textbf{Model}& \textbf{RMSE}& \textbf{MAE}& \textbf{MAPE}
& \textbf{RRSE}\\
\hline
FBProphet&21.2&14.4&34.7&0.295  \\
ARIMA& 21.2&14.1 &30.5&0.308 \\
LSTM& 20.8& 13.2&22.7&0.292 \\
CNN& 22.4&14.3 &26.3&0.312 \\
\hline
\end{tabular}

\end{center}
\end{table}

We can see from Fig. 7 that there isn't much difference in RMSE and MAE values. However, this difference widened for MAPE values, which show that ARIMA and FBProphet neglected to anticipate the PM2.5 levels consist of smaller peaks.

Since LSTM is giving the best results, it is important to know which activation function performed better in forecasting. The average results for the 'tanh' and 'relu' activation function used in LSTM to forecast PM2.5 levels in all 12 stations are given in Table IV. The results clearly state that LSTM performance improved with the nonlinearity of relu [21].

\begin{figure}[htbp]
\centerline{\includegraphics[width=8cm, height=4.5cm]{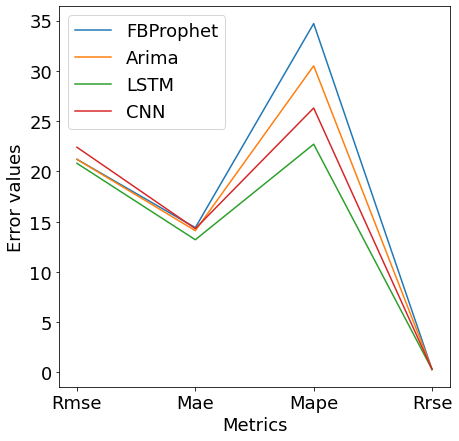}}
\caption{Line Chart of Averaged predicted results}
\end{figure}
\begin{figure}[htbp]
\centerline{\includegraphics[width=8cm, height=3.5cm]{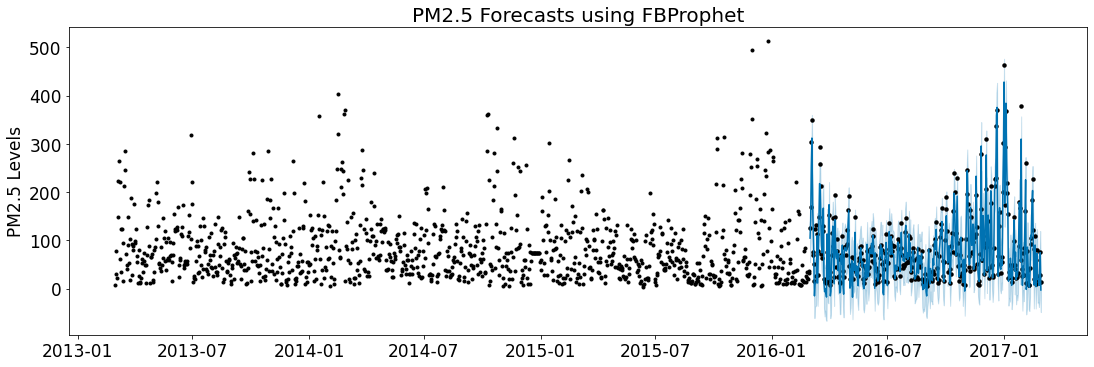}}
\centerline{\includegraphics[width=8cm, height=3.5cm]{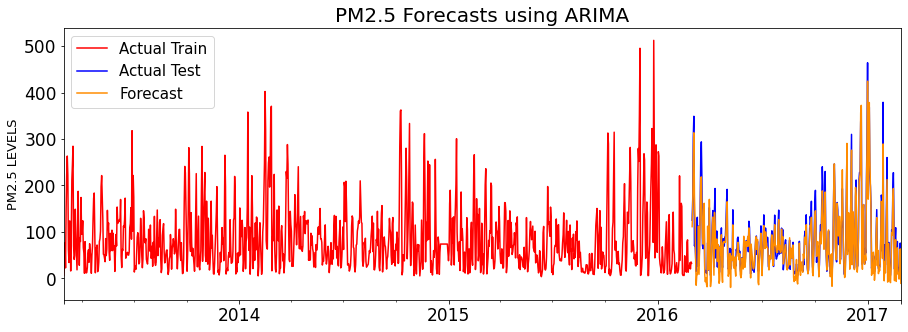}}
\centerline{\includegraphics[width=8cm, height=3.5cm]{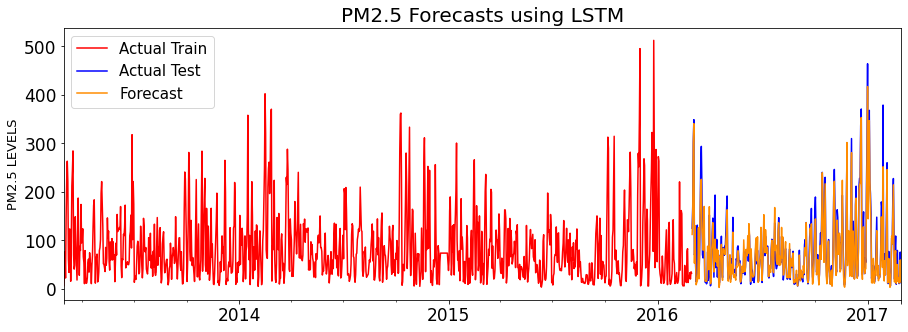}}
\centerline{\includegraphics[width=8cm, height=3.5cm]{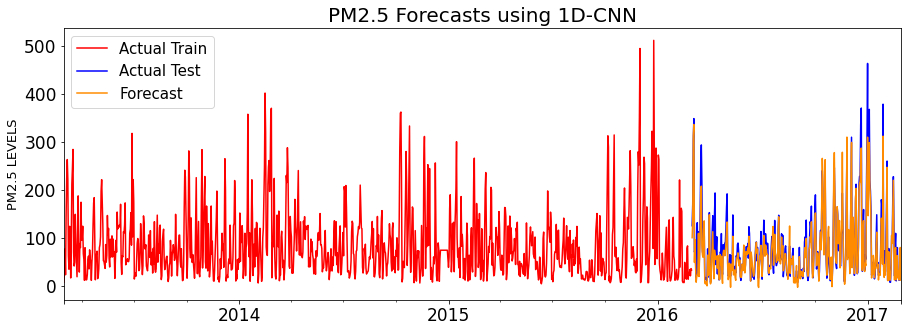}}
\caption{Actual and forecasted values subplots of FBProphet, ARIMA, LSTM, 1D-CNN on Aotizhongxin station using full dataset.}
\end{figure}

\begin{table}[htbp]
\caption{Averaged forecast results using tanh and relu activation on LSTM }
\begin{center}
\begin{tabular}{|c|c|c|c|c|}
\hline
\textbf{Activation}& \textbf{RMSE}& \textbf{MAE}& \textbf{MAPE}
& \textbf{RRSE}\\
\hline
 tanh&22.9& 14.4& 25.9& 0.320 \\
 relu&21.2& 13.3& 22.9& 0.297 \\
\hline
\end{tabular}
\end{center}
\end{table}

\begin{figure}[htbp]
\centerline{\includegraphics[width=8cm, height=3.5cm]{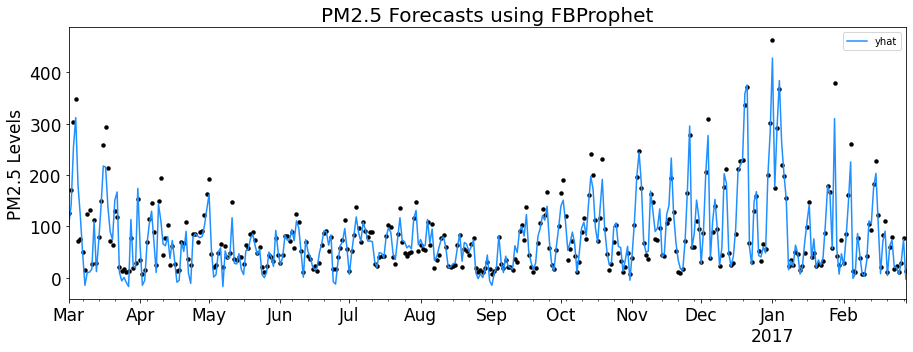}}
\centerline{\includegraphics[width=8cm, height=3.5cm]{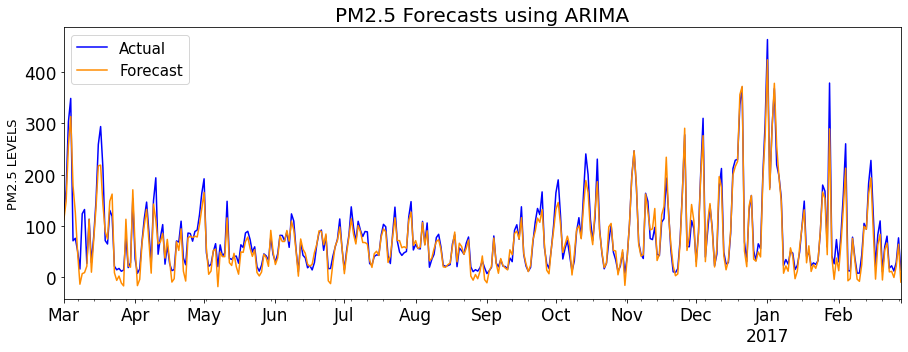}}
\centerline{\includegraphics[width=8cm, height=3.5cm]{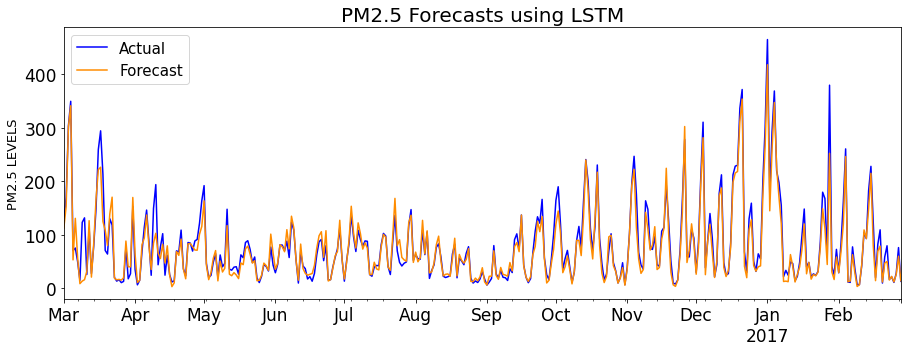}}
\centerline{\includegraphics[width=8cm, height=3.5cm]{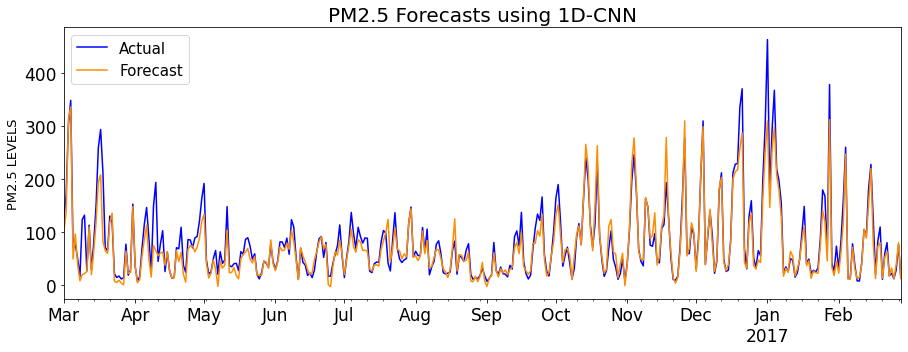}}
\caption{Actual and forecasted values subplots of FBProphet, ARIMA, LSTM, 1D-CNN on Aotizhongxin station using test data only.}
\end{figure}

The fitting of PM2.5 actual and forecasted levels by all four models is featured for visualization. For this analysis, we only choose the Aotizhongxin station. The results by all four models, FBProphet, ARIMA, LSTM, and CNN, are shown in Fig. 8. For the FBProphet model, the actual data pictured as black points. The blue line represents the forecasted data with lower and upper confidence intervals in a light blue region, while for all other models, the training and test data are represented as a red line and green line, respectively.
The forecast is an orange line overlapping the green line of test data. From the plots, we can easily analyze that, except for CNN, all models covered the peaks well, proving the RMSE value of CNN is slightly lower than all other models. 

Subsequently, we can analyze that the LSTM and CNN covered the small peaks well compared to Arima and FBProphet, which suggests the low MAPE error in LSTM, CNN, and high MAPE error in ARIMA and FBProphet.

For clear visualization, the test data plots from March 1, 2016, to February 28, 2017, are also picturized, as shown in Fig. 9.

\section{Conclusion and future work}
The Heart and respiratory-related diseases like COPD, stroke, lung cancer are correlated with air pollution. Not just humans, it also affects other living organisms and damages the natural environment. This paper provided an analysis and prediction study of the PM2.5 levels on 12 station sites using four models; ARIMA, FBProphet, LSTM, and CNN. For most of the stations, LSTM performed better than all other models across RRSE, RMSE, MAPE, and MAE evaluation metrics. We had also shown that LSTM with relu activation function performed better than tanh, which opens up the possibility of spending more time analyzing hyper optimization techniques. The trend analysis shows that ARIMA and FBProphet did not predict smaller values correctly, resulting in high MAPE error. The adopted methods achieved good results. However, it restricts our examination to the model's adequacy, which can be additionally improved by considering an ensemble of various forecast models. The obtained forecasting results can be improved by taking feature engineering and better hyperparameter optimization into account, which will be a part of the future work.

\vspace{12pt}

\end{document}